# A NOVEL METRIC APPROACH EVALUATION FOR THE SPATIAL ENHANCEMENT OF PAN-SHARPENED IMAGES

Firouz Abdullah Al-Wassai[1] and Dr. N.V. Kalyankar[2]

[1]Department of Computer Science, (SRTMU), Nanded, India
fairozwaseai@yahoo.com
[2]Principal, Yeshwant Mahavidyala College, Nanded, India
drkalyankarnv@yahoo.com

## ABSTRACT

Various and different methods can be used to produce high-resolution multispectral images from high-resolution panchromatic image (PAN) and low-resolution multispectral images (MS), mostly on the pixel level. The Quality of image fusion is an essential determinant of the value of processing images fusion for many applications. Spatial and spectral qualities are the two important indexes that used to evaluate the quality of any fused image. However, the jury is still out of fused image's benefits if it compared with its original images. In addition, there is a lack of measures for assessing the objective quality of the spatial resolution for the fusion methods. So, an objective quality of the spatial resolution assessment for fusion images is required. Therefore, this paper describes a new approach proposed to estimate the spatial resolution improve by High Past Division Index (HPDI) upon calculating the spatial-frequency of the edge regions of the image and it deals with a comparison of various analytical techniques for evaluating the Spatial quality, and estimating the colour distortion added by image fusion including: MG, SG, FCC, SD, En, SNR, CC and NRMSE. In addition, this paper devotes to concentrate on the comparison of various image fusion techniques based on pixel and feature fusion technique.

## 1. INTRODUCTION

The Quality of image fusion is an essential determinant of the value of processing images fusion for many applications. Spatial and spectral qualities are the two important indexes that used to evaluate the quality of any fused image. Generally, one aims to preserve as much source information as possible in the fused image with the expectation that performance with the fused image will be better than, or at least as good as, performance with the source images [1]. Several authors describe different spatial and spectral quality analysis techniques of the fused images. Some of them enable subjective, the others objective, numerical definition of spatial or spectral quality of the fused data [2-5]. The evaluation of the spatial quality of the pan-sharpened images is equally important since the goal is to retain the high spatial resolution of the PAN image. A survey of the pan sharpening literature revealed there were very few papers that evaluated the spatial quality of the pan-sharpened imagery [6]. However, the jury is still out on the benefits of a fused image compared to its original images. There is also a lack of measures for assessing the objective quality of the spatial resolution of the fusion methods. As a result of that, an objective quality of the spatial resolution assessment for fusion images is required.

This study presented a new approach to assess the spatial quality of a fused image based on HPDI, depends upon the spatial-frequency of the edge regions of the image and comparing it with other methods as [27, 28, 33]. In addition, many spectral quality metrics, to compare the properties of fused images and their ability to preserve the similarity with respect to the MS image while incorporating the spatial resolution of the PAN image, should increase the spectral fidelity while retaining the spatial resolution of the PAN. In addition, this study focuses on cambering that the best methods based on pixel fusion techniques (see section 2) are those with the fallowing feature fusion techniques: Segment Fusion (SF), Principal Component Analysis based Feature Fusion (PCA) and Edge Fusion (EF) in [7].

The paper organized as follows .Section 2 gives the image fusion techniques; Section 3 includes the quality of evaluation of the fused images; Section 4 covers the experimental results and analysis then subsequently followed by the conclusion.

## 2. IMAGE FUSION TECHNIQUES

Image fusion techniques can be divided into three levels, namely: pixel level, feature level and decision level of representation [8-10]. The image fusion techniques based on pixel can be grouped into several techniques depending on the tools or the processing methods for image fusion procedure summarized as fallow:

1) Arithmetic Combination techniques: such as Bovey Transform (BT) [11-13]; Color Normalized Transformation (CN) [14, 15]; Multiplicative Method (MLT) [17, 18].
2) Component Substitution fusion techniques: such as IHS, HSV, HLS and YIQ in [19].
3) Frequency Filtering Methods :such as in [20] High-Pass Filter Additive Method (HPFA) , High – Frequency- Addition Method (HFA) , High Frequency Modulation Method (HFM) and The Wavelet transform-based fusion method (WT).
4) Statistical Methods: such as in [21] Local Mean Matching (LMM), Local Mean and Variance Matching (LMVM), Regression variable substitution (RVS), and Local Correlation Modeling (LCM).

All the above techniques employed in our previous studies [19-21]. Therefore, the best method for each group selected in this study as the fallowing: (HFA), (HFM) [20], (RVS) [21] and the IHS method by [22].

To explain the algorithms through this study, Pixels should have the same spatial resolution from two different sources that are manipulated to obtain the resultant image. Here, The PAN image have a different spatial resolution from that of the MS image. Therefore, re-sampling of MS image to the spatial resolution of PAN is an essential step in some fusion methods to bring the MS image to the same size of PAN, thus the re-sampled MS image will be noted by $M_k$ that represents the set of DN of band k in the re-sampled MS image.

## 3. QUALITY EVALUATION OF THE FUSED IMAGES

This section describes the various spatial and spectral quality metrics used to evaluate them. The spectral fidelity of the fused images with respect to the MS images is described. When analyzing the spectral quality of the fused images we compare spectral characteristics of images obtained from the different methods, with the spectral characteristics of re-sampled MS images. Since the goal is to preserve the radiometry of the original MS images, any metric used must measure the amount of change in DN values in the pan-sharpened image $F_k$ compared to the original image $M_k$. Also, In order to evaluate the spatial properties of the fused images, a PAN image and intensity image of the fused image have to be compared since the goal is to retain the high spatial resolution of the PAN image. In the following $F_k$ , $M_k$ are the measurements of each the brightness values pixels of the result image and the original MS image of bandk, $\overline{M}_k$ and $\overline{F}_k$ are the mean brightness values of both images and are of size $n * m$. BV is the brightness value of image data $\overline{M}_k$ and $\overline{F}_k$.

## 3.1 SPECTRAL QUALITY METRICS:

1) **Standard Deviation (SD)**

   The SD, which is the square root of variance, reflects the spread in the data. Thus, a high contrast image will have a larger variance, and a low contrast image will have a low variance. It indicates the closeness of the fused image to the original MS image at a pixel level. The ideal value is zero.

   $$\sigma = \sqrt{\frac{\sum_{i=1}^{m}\sum_{j=1}^{n}(BV(n,m)-\mu)^2}{m \times n}} \quad (1)$$

2) **Entropy (En)**

   The En of an image is a measure of information content but has not been used to assess the effects of information change in fused images. En reflects the capacity of the information carried by images. The larger En means that high information in the image [6]. By applying Shannon's entropy in evaluation the information content of an image, the formula is modified as [23]:

   $$En = -\sum_{i=0}^{255} P(i) \log_2 P(i) \quad (2)$$

   Where P(i) is the ratio of the number of the pixels with gray value equal to $i$ over the total number of the pixels.

3) **Signal-to Noise Ratio (SNR)**

   The signal is the information content of the data of MS image $M_k$, while the merging $F_k$ can cause the noise, as error that is added to the signal. The $RMS_k$ of the SNR can be used to calculate the $SNR_k$, given by [24]:

   $$SNR_k = \sqrt{\frac{\sum_i^n \sum_j^m (F_k(i,j))^2}{\sum_i^n \sum_j^m (F_k(i,j)-M_k(i,j))^2}} \quad (3)$$

4) **Correlation Coefficient (CC)**

   The CC measures the closeness or similarity between two images. It can vary between –1 to +1. A value close to +1 indicates that the two images are very similar, while a value close to –1 indicates that they are highly dissimilar. The formula to compute the correlation between $F_k$, $M_k$:

   $$CC = \frac{\sum_i^n \sum_j^m (F_k(i,j)-\overline{F}_k)(M_k(i,j)-\overline{M}_k)}{\sqrt{\sum_i^n \sum_j^m (F_k(i,j)-\overline{F}_k)^2}\sqrt{\sum_i^n \sum_j^m (M_k(i,j)-\overline{M}_k)^2}} \quad (4)$$

   Since the pan-sharpened image larger (more pixels) than the original MS image it is not possible to compute the CC or apply any other mathematical operation between them. Thus, the up-sampled MS image $M_k$ is used for this comparison.

5) **Normalization Root Mean Square Error (NRMSE)**

   The NRMSE used in order to assess the effects of information changing for the fused image. When level of information loss can be expressed as a function of the original MS pixel $M_k$ and the fused pixel $F_k$, by using the NRMSE between $M_k$ and $F_k$ images in band k. The Normalized Root- Mean-Square Error $NRMSE_k$ between $F_k$ and $M_k$ is a point analysis in multispectral space representing the amount of change the original MS pixel and the corresponding output pixels using the following equation [27]:

$$NRMSE_k = \sqrt{\frac{1}{nm*255^2}\sum_i^n \sum_j^m (F_k(i,j) - M_k(i,j))^2} \quad (5)$$

6) **The Histogram Analysis**

The histograms of the multispectral original MS and the fused bands must be evaluated [4]. If the spectral information preserved in the fused image, its histogram will closely resemble the histogram of the MS image. The analysis of histogram deals with the brightness value histograms of all RGB-color bands, and L-component of the resample MS image and the fused A greater difference of the shape of the corresponding histograms represents a greater spectral change [31].

## 3.2 SPATIAL QUALITY METRICS

1) **Mean Grades (MG)**

MG has been used as a measure of image sharpness by [27, 28]. The gradient at any pixel is the derivative of the DN values of neighboring pixels. Generally, sharper images have higher gradient values. Thus, any image fusion method should result in increased gradient values because this process makes the images sharper compared to the low-resolution image. The calculation formula is [6]:

$$\bar{G} = \frac{1}{(m-1)(n-1)}\sum_{i=1}^{m-1}\sum_{j=1}^{n-1}\sqrt{\frac{\Delta I_x^2 + \Delta I_y^2}{2}} \quad (6)$$

Where

$$\Delta I_x = f(i+1,j) - f(i,j)$$
$$\Delta I_y = f(i,j+1) - f(i,j) \quad (7)$$

Where $\Delta I_x$ and $\Delta I_y$ are the horizontal and vertical gradients per pixel of the image fused $f(i,j)$. generally, the larger $\bar{G}$, the more the hierarchy, and the more definite the fused image.

2) **Soble Grades (SG)**

This approach developed in this study by used the Soble operator. That by computes discrete gradient in the horizontal and vertical directions at the pixel location $i,j$ of an image $f(i,j)$. The Soble operator was the most popular edge detection operator until the development of edge detection techniques with a theoretical basis. It proved popular because it gave a better performance contemporaneous edge detection operator than other such as the Prewitt operator [30]. For this, which is clearly more costly to evaluate, the orthogonal components of gradient as the following [31]:

$$Gx = \{f(i-1,j+1) + 2f(i-1,j) + f(i-1,j-1)\} - \{f(i+1,j+1) + 2f(i+1,j) + f(i+1,j-1)\}$$

And

$$Gy = \{f(i-1,j+1) + 2f(i,j+1) + f(i+1,j+1)\} - \{f(i-1,j-1) + 2f(i,j-1) + f(i+1,j-1)\} \quad (8)$$

It can be seen that the Soble operator is equivalent to simultaneous application of the templates as the following [32]:

$$G_x = \begin{bmatrix} 1 & 2 & 1 \\ 0 & 0 & 0 \\ -1 & -2 & -1 \end{bmatrix} \quad G_y = \begin{bmatrix} -1 & 0 & 1 \\ -2 & 0 & 2 \\ -1 & 0 & 1 \end{bmatrix} \quad (9)$$

Then the discrete gradient $G$ of an image $f(i,j)$ is given by

$$\bar{G} = \frac{1}{(m-1)(n-1)} \sum_{i=1}^{(m-1)} \sum_{j=1}^{(n-1)} \sqrt{\frac{G_x^2 + G_y^2}{2}} \quad (10)$$

Where $G_x$ and $G_y$ are the horizontal and vertical gradients per pixel. Generally, the larger values for $\bar{G}$, the more the hierarchy and the more definite the fused image.

### 3) Filtered Correlation Coefficients (FCC)

This approach was introduced [33]. In the Zhou's approach, the correlation coefficients between the high-pass filtered fused PAN and TM images and the high-pass filtered PAN image are taken as an index of the spatial quality. The high-pass filter is known as a Laplacian filter as illustrated in eq. (11)

$$\text{mask} = \begin{bmatrix} -1 & -1 & -1 \\ -1 & 8 & -1 \\ -1 & -1 & -1 \end{bmatrix} \quad (11)$$

However, the magnitude of the edges does not necessarily have to coincide, which is the reason why Zhou et al proposed to look at their correlation coefficients [33]. So, in this method the average correlation coefficient of the faltered PAN image and all faltered bands is calculated to obtain FCC. An FCC value close to one indicates high spatial quality.

### 4) HPDI a New Scheme Of Spatial Evaluation Quality

To explain the new proposed technique of HPDI to evaluation the quality of the spatial resolution specifying the edges in the image by using the Laplacian filter (eq.11). The Laplacian filtered PAN image is taken as an index of the spatial quality to measure the amount of edge information from the PAN image is transferred into the fused images. The deviation index between the high pass filtered $P$ and the fused $F_k$ images would indicate how much spatial information from the PAN image has been incorporated into the $MS$ image to obtain HPDI as follows:

$$\text{HPDI}_k = \frac{1}{nm} \sum_i^n \sum_j^m \frac{|F_k(i,j) - P(i,j)|}{P(i,j)} \quad (12)$$

The larger value HPDI the better image quality. Indicates that the fusion result it has a high spatial resolution quality of the image.

## 4. EXPERIMENTAL RESULTS

The above assessment techniques are tested on fusion of Indian IRS-1C PAN of the 5.8- m resolution panchromatic band and the Landsat TM the red (0.63 - 0.69 µm), green (0.52 - 0.60 µm) and blue (0.45 - 0.52 µm) bands of the 30 m resolution multispectral image were used in this work. Fig.1 shows the IRS-1C PAN and multispectral TM images. Hence, this work is an attempt to study the quality of the images fused from different sensors with various characteristics. The size of the PAN is 600 * 525 pixels at 6 bits per pixel and the size of the original multispectral is 120 * 105 pixels at 8 bits per pixel, but this is up-sampled by nearest neighbor to same size the PAN image. The pairs of images were geometrically registered to each other. The HFA, HFM, HIS, RVS, PCA, EF, and SF methods are employed to fuse IRS-C PAN and TM multi-spectral images.

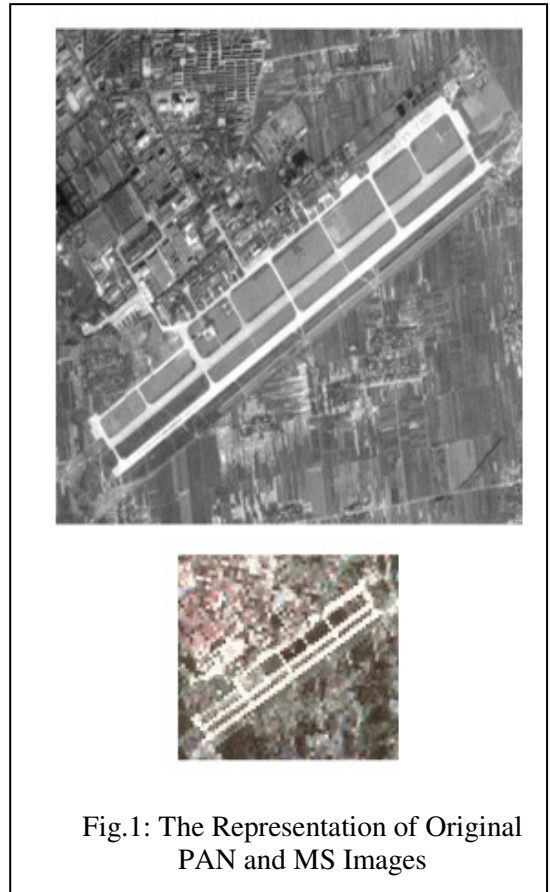

Fig.1: The Representation of Original PAN and MS Images

### 4.1. Spectral Quality Metrics Results

From table1 and Fig. 2 shows those parameters for the fused images using various methods. It can be seen that from Fig. 2a and table1 the SD results of the fused images remains constant for all methods except the IHS. According to the computation results En in table1, the increased En indicates the change in quantity of information content for spectral resolution through the merging. From table1 and Fig.2b, it is obvious that En of the fused images have been changed when compared to the original MS except the PCA. In Fig.2c and table1 the maximum correlation values was for PCA. In Fig.2d and table1 the maximum results of SNR were with the SF, and HFA. Results SNR and NRMSE appear changing significantly. It can be observed from table1 with the diagram Fig. 2d & Fig. 2e for results SNR and NRMSE of the fused image, the SF and HFA methods gives the best results with respect to the other methods. Means that this method maintains most of information spectral content of the original MS data set which gets the same values presented the lowest value of the NRMSE as well as the high of the CC and SNR. Hence, the SF and HFA fused images for preservation of the spectral resolution original MS image much better techniques than the other methods.

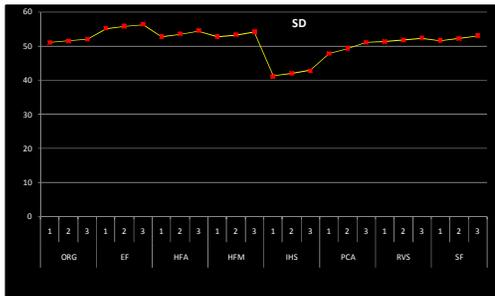
Fig. 2a: Chart Representation of SD

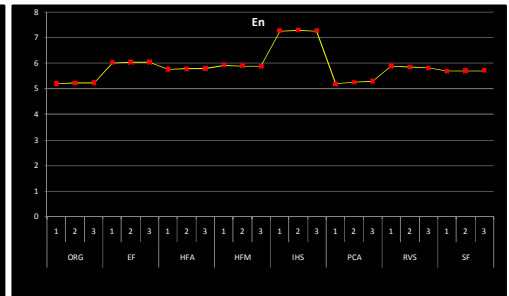
Fig. 2b: Chart Representation of En

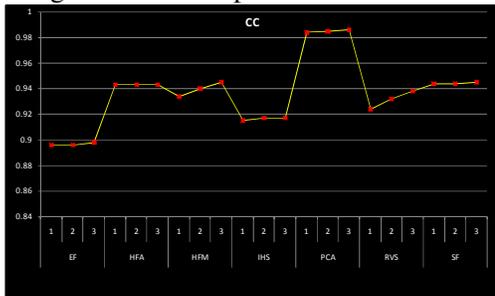
Fig.2c: Chart Representation of CC

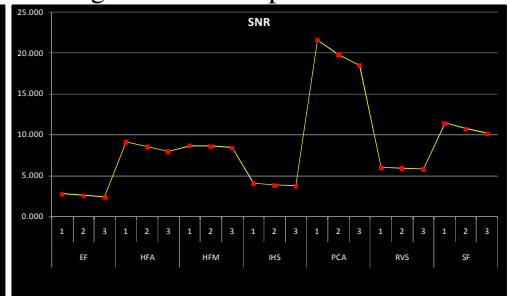
Fig. 2d: Chart Representation of SNR

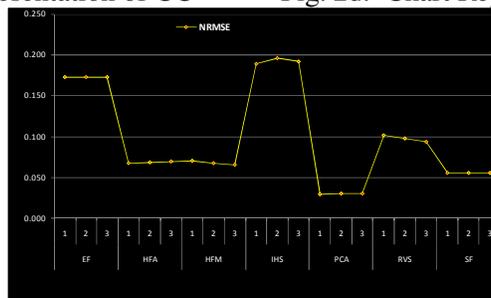
Fig. 2e: Chart Representation of NRMSE
Fig. 2: Chart Representation of SD, En, CC, SNR, NRMSE of Fused Images

Table 1: The Spectral Quality Metrics Results for the Original MS and Fused Image Methods

| Method | Band | SD | En | CC | SNR | NRMSE |
|---|---|---|---|---|---|---|
| ORG | 1 | 51.018 | 5.2093 | -- | --- | ---- |
| | 2 | 51.477 | 5.2263 | --- | --- | ---- |
| | 3 | 51.983 | 5.2326 | --- | --- | --- |
| EF | 1 | 55.184 | 6.0196 | 0.896 | 2.742 | 0.173 |
| | 2 | 55.792 | 6.0415 | 0.896 | 2.546 | 0.173 |
| | 3 | 56.308 | 6.0423 | 0.898 | 2.359 | 0.173 |
| HFA | 1 | 52.793 | 5.7651 | 0.943 | 9.107 | 0.068 |
| | 2 | 53.57 | 5.7833 | 0.943 | 8.518 | 0.069 |
| | 3 | 54.498 | 5.7915 | 0.943 | 7.946 | 0.070 |
| HFM | 1 | 52.76 | 5.9259 | 0.934 | 8.660 | 0.071 |
| | 2 | 53.343 | 5.8979 | 0.94 | 8.581 | 0.068 |
| | 3 | 54.136 | 5.8721 | 0.945 | 8.388 | 0.066 |
| IHS | 1 | 41.164 | 7.264 | 0.915 | 4.063 | 0.189 |
| | 2 | 41.986 | 7.293 | 0.917 | 3.801 | 0.196 |
| | 3 | 42.709 | 7.264 | 0.917 | 3.690 | 0.192 |
| PCA | 1 | 47.875 | 5.1968 | 0.984 | 21.611 | 0.030 |
| | 2 | 49.313 | 5.2485 | 0.985 | 19.835 | 0.031 |
| | 3 | 51.092 | 5.2941 | 0.986 | 18.515 | 0.031 |
| RVS | 1 | 51.323 | 5.8841 | 0.924 | 5.936 | 0.102 |
| | 2 | 51.769 | 5.8475 | 0.932 | 5.887 | 0.098 |
| | 3 | 52.374 | 5.8166 | 0.938 | 5.800 | 0.094 |
| SF | 1 | 51.603 | 5.687 | 0.944 | 11.422 | 0.056 |
| | 2 | 52.207 | 5.7047 | 0.944 | 10.732 | 0.056 |
| | 3 | 53.028 | 5.7123 | 0.945 | 10.144 | 0.056 |

The spectral distortion introduced by the fusion can be analyzed the histogram for all RGB color bands and L-component that appears changing significantly. Fig.2 noted that the matching for R &G color bands between the original MS with the fused images. Many of the image fusion methods examined in this study and the best matching for the intensity values between the original MS image and the fused image for each of the R&G color bands obtained by SF. There are also matching for the B color band in Fig.3 and L-component in Fig.3 except at the values of intensity that ranging in value 253 to 255 not appear the values intensity of the original image whereas highlight the values of intensity of the merged images clearly in the Fig.2 & Fig.3. That does not mean its conflicting values or the spectral resolution if we know that the PAN band (0.50 - 0.75 µm) does not spectrally overlap the blue band of the MS (0.45 - 0.52 µm). Means that during the process of merging been added intensity values found in the PAN image and there have been no in the original MS image which are subject to short wavelengths affected by many factors during the transfer and There can be no to talk about these factors in this context. Most researchers histogram match the PAN band to each MS band before merging them and substituting the high frequency coefficients of the PAN image in place of the MS image's coefficients such as HIS &PCA methods . However, they have been found where the radiometric normalization as EF &PCA methods is left out Fig. 2 &3. Also, by analyzing the histogram of the Fig. 3 for the fused image, we found that the values of intensity are more significantly when values of 255 for the G&B color bands of the original MS image. The extremism in the Fig. 3 for the intensity of luminosity disappeared.

Generally, the best method through the previous analysis of the Fig.2 and Fig.3 to preservation of the maximum spectral characteristics as possible to the original image for each RGB band and L-component was with SF method. Because the edges are affected more than homogenous regions through the process of merge by moving spatial details to the multispectral MS image and consequently affect on its features and that showed in the image after the merged.

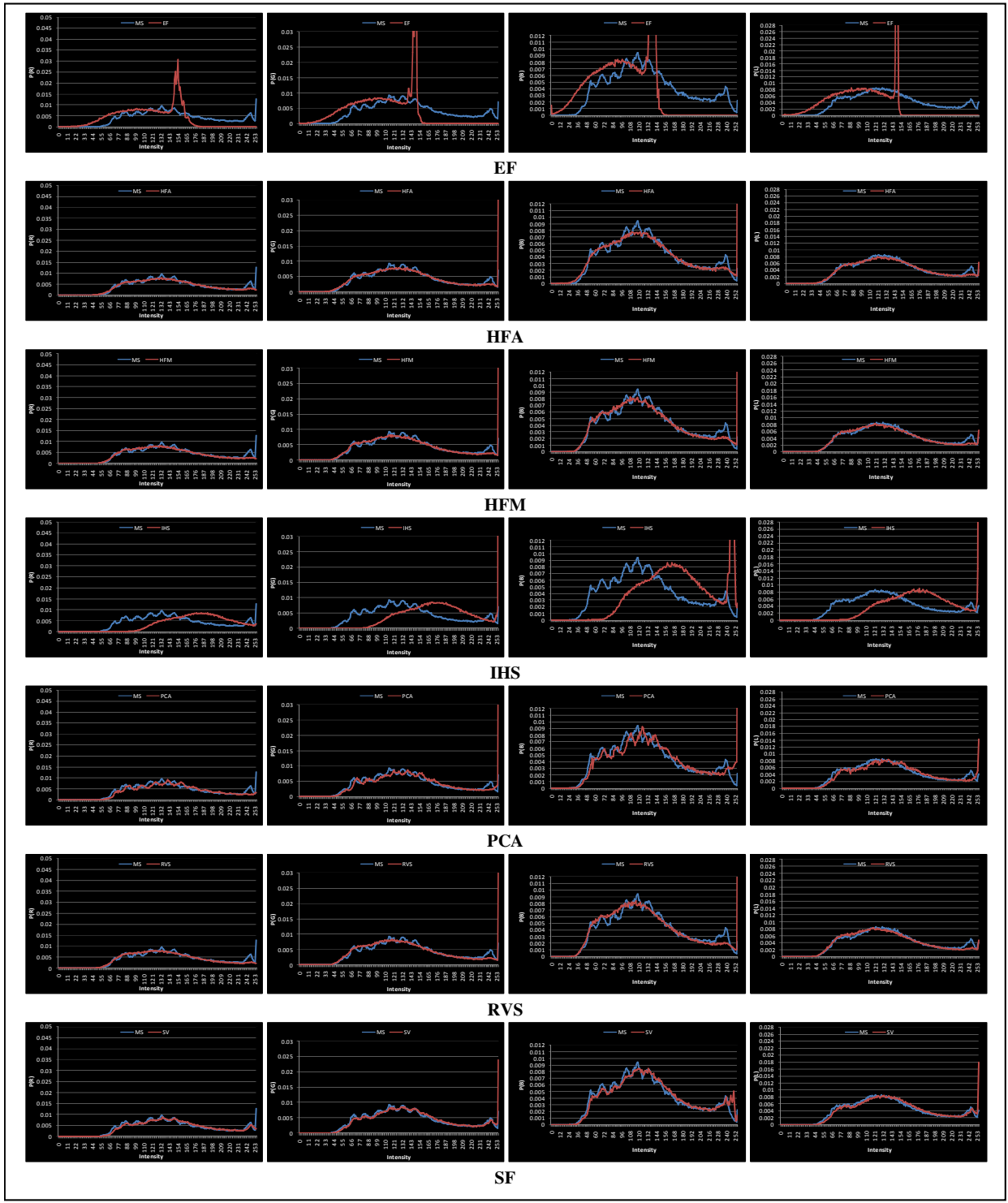

Fig.3: Histogram Analysis for All RGB-Color Band and L-Component of Fused Images with MS Image

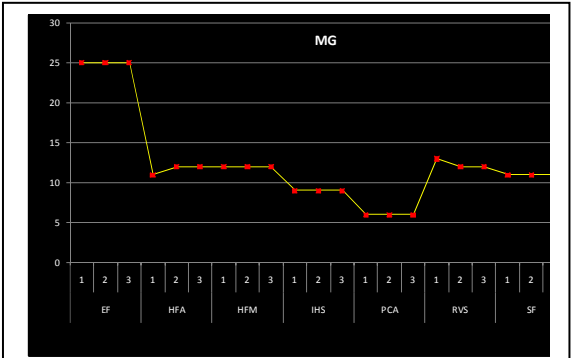

Fig. 4a: Chart Representation of MG

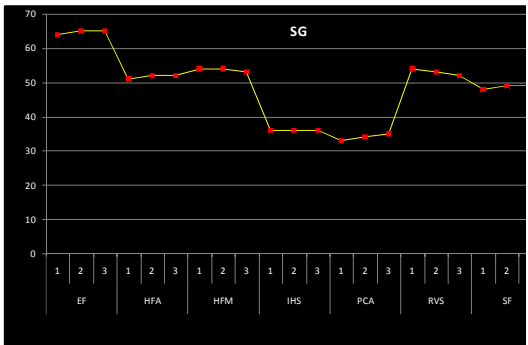

Fig. 4b: Chart Representation of SG

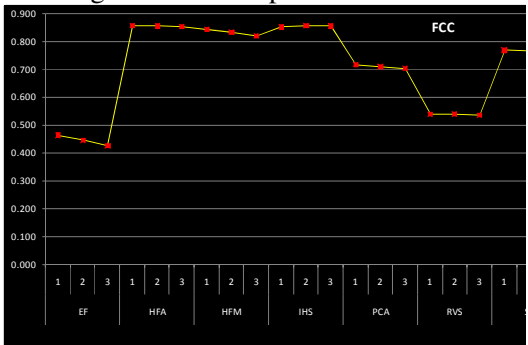

Fig. 4c: Chart Representation of FCC

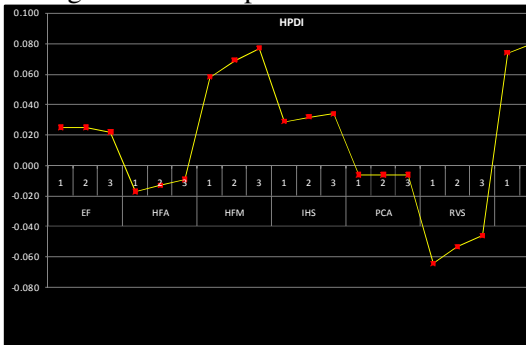

Fig. 4d: Chart Representation of HPDI

Fig. 4: Chart Representation of MG, SG, FCC & HPDI of Fused Images

## 4.2 Spatial Quality Metrics Results:

Table 2 and Fig. 5 show the result of the fused images using various methods. It is clearly that the seven fusion methods are capable of improving the spatial resolution with respect to the original MS image. From table2 and Fig. 4 shows those parameters for the fused images using various methods. It can be seen that from Fig. 4a and table2 the MG results of the fused images increase the spatial resolution for all methods except the PCA and IHS. Also, in Fig.4a the maximum gradient for SG was 64 edge but for MG in table2 and Fig.4b the maximum gradient was 25 edge means that the SG it gave, overall, a better performance than MG to edge detection. In addition, the SG appears the results of the fused images increase the gradient for all methods except the PCA means that the decreasing in gradient that it dose not enhance the spatial quality. The maximum results of MG and SG for sharpen image methods was for the EF but the nearest to the PAN it was SF has the same results approximately. However, when comparing them to the PAN it can be seen that the SF close to the Result of the PAN. Other means the SF added the details of the PAN image to the MS image as well as the maximum preservation of the spatial resolution of the PAN.

According to the computation results, FCC in table2 and Fig.4c the increase FCC indicates the amount of edge information from the PAN image transferred into the fused images in quantity of spatial resolution through the merging. The maximum results of FCC From table2 and Fig.4c were for the HFA, HFM and SF. The purposed approach of HPDI as the spatial quality metric is more important than the other spatial quality matrices to distinguish the best spatial enhancement through the merging. Also, the analytical technique of HPDI is much more useful for measuring the spatial enhancement corresponding to the Pan image than the other methods since the FCC or SG and MG gave the same results for some methods; but the HPDI gave the smallest different ratio between those methods. It can be observed that from Fig.4d and table2 the maximum results of the purpose approach HPDI it were with the SF followed HFM methods.

Table 2: The Spatial Quality Results of Fused Images

| Method | Band | MG | SG | HPDI | FCC |
|---|---|---|---|---|---|
| **EF** | 1 | 25 | 64 | 0.025 | 0.464 |
| | 2 | 25 | 65 | 0.025 | 0.446 |
| | 3 | 25 | 65 | 0.022 | 0.426 |
| **HFA** | 1 | 11 | 51 | -0.017 | 0.857 |
| | 2 | 12 | 52 | -0.013 | 0.856 |
| | 3 | 12 | 52 | -0.009 | 0.854 |
| **HFM** | 1 | 12 | 54 | 0.058 | 0.845 |
| | 2 | 12 | 54 | 0.069 | 0.834 |
| | 3 | 12 | 53 | 0.077 | 0.821 |
| **IHS** | 1 | 9 | 36 | 0.029 | 0.853 |
| | 2 | 9 | 36 | 0.032 | 0.857 |
| | 3 | 9 | 36 | 0.034 | 0.856 |
| **PCA** | 1 | 6 | 33 | -0.006 | 0.718 |
| | 2 | 6 | 34 | -0.006 | 0.710 |
| | 3 | 6 | 35 | -0.006 | 0.704 |
| **RVS** | 1 | 13 | 54 | -0.064 | 0.539 |
| | 2 | 12 | 53 | -0.053 | 0.540 |
| | 3 | 12 | 52 | -0.046 | 0.536 |
| **SF** | 1 | 11 | 48 | 0.074 | 0.769 |
| | 2 | 11 | 49 | 0.080 | 0.766 |
| | 3 | 11 | 49 | 0.080 | 0.761 |
| **PAN** | 1 | 10 | 42 | -- | -- |

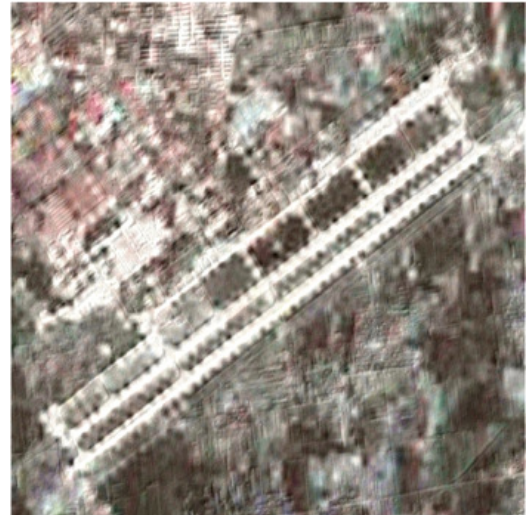

Fig.5a: HFA

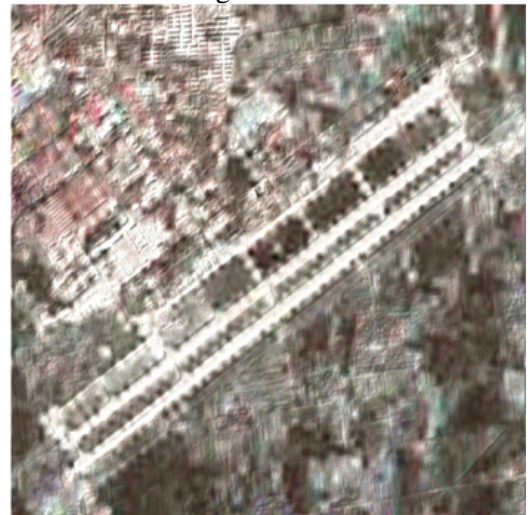

Fig.5b: HFM

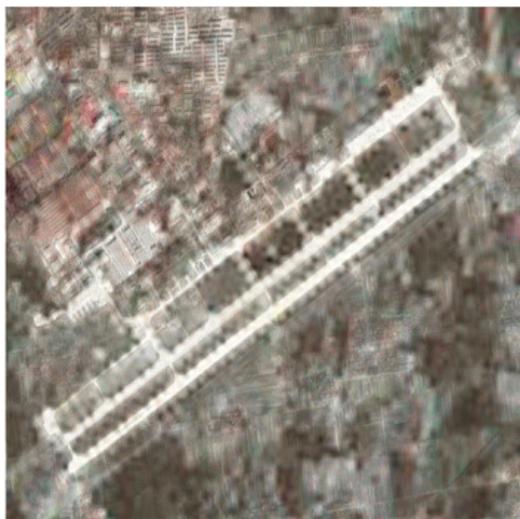      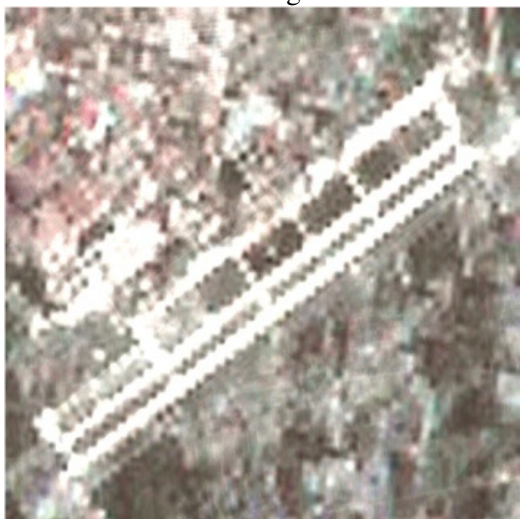

Fig.5c: IHS                         Fig.5d: PCA

Fig.5: The Representation of Fused Images

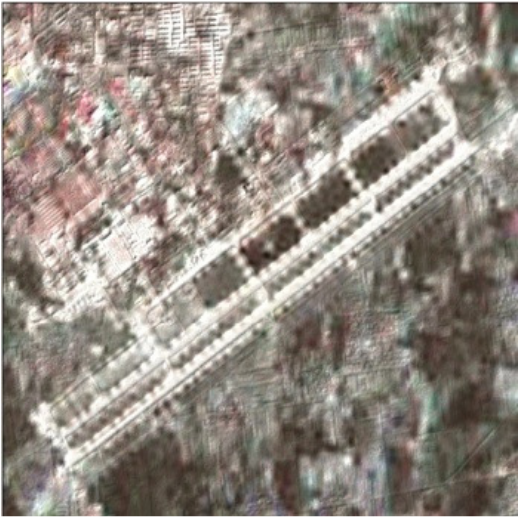
Fig.5e: RVS

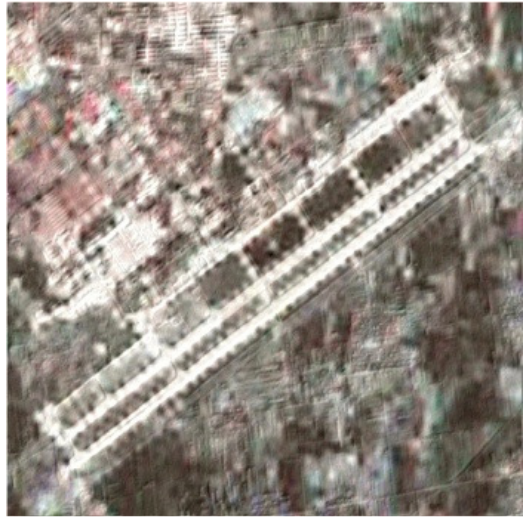
Fig.5f: SF

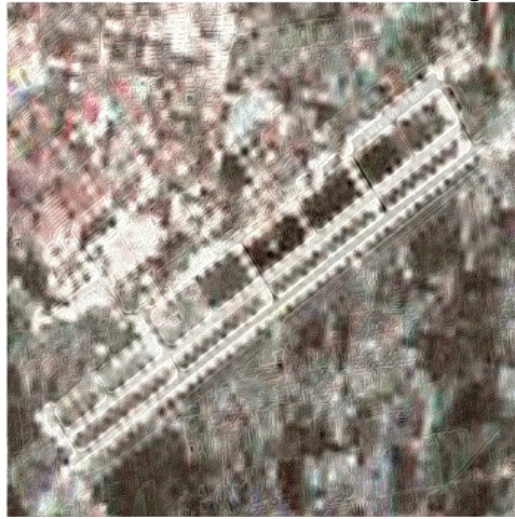
Fig.5g: EF
Continue Fig.5: The Representation of Fused Images

## 5. CONCLUSION

This study proposed a new measure to test the efficiency of spatial resolution of fusion images applied to a number of methods of merge images. These methods have obtained the best results in previous studies and some of them depend on the pixel level fusion including HFA, HFM, IHS and RVS methods while the other methods based on features level fusion like PCA, EF and SF method. Results of the study show the importance to propose a new HPDI as a criterion to measure the quality evaluation for the spatial resolution of the fused images in which the results showed the effectiveness of high efficiency when compared with the other criterion methods for measurement such as the FCC. The proposed analytical technique of HPDI is much more useful for measuring the spatial enhancement of fused image corresponding to the spatial resolution of the PAN image than the other methods, since the FCC or SG and MG gave the same results for some methods; but the HPDI gave the smallest different ratio between those methods, therefore, it is strongly recommended to use HPDI for measuring the spatial enhancement of fused image with PAN image because of its mathematical and more precision as quality indicator.

Experimental results with spatial and spectral quality matrices evaluation further show that the SF technique based on feature level fusion maintains the spectral integrity for MS image as well as improved as much as possible the spatial quality of the PAN image. The use of the SF based fusion technique is strongly recommended if the goal of merging is to achieve the best representation of the spectral information of multispectral image and the spatial details of a high-resolution panchromatic image. Because it is based on Component Substitution fusion techniques coupled with a spatial domain filtering. It utilizes the statistical variable between the brightness values of the image bands to adjust the contribution of individual bands to the fusion results to reduce the color distortion. The analytical technique of SG is much more useful for measuring gradient than MG as the MG gave the smallest gradient results.

**Authors**

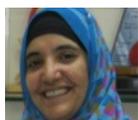    Firouz Abdullah Al-Wassai. Received the B.Sc. degree in physics from University of Sana'a, Yemen in 1993; the M. Sc. degree from Bagdad University, Iraq in 2003. Currently, she is Ph. D. scholar in computer Science at department of computer science (S.R.T.M.U), Nanded, India.

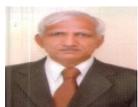    Dr. N.V. Kalyankar,He is a Principal of Yeshwant Mahvidyalaya, Nanded(India) completed M.Sc.(Physics) from Dr. B.A.M.U, Aurangabad. In 1980 he joined as a leturer in department of physics at Yeshwant Mahavidyalaya, Nanded. In 1984 he completed his DHE. He completed his Ph.D. from Dr.B.A.M.U, Aurangabad in 1995. From 2003 he is working as a Principal to till date in Yeshwant Mahavidyalaya, Nanded. He is also research guide for Physics and Computer Science in S.R.T.M.U, Nanded. 03 research students are successfully awarded Ph.D in Computer Science under his guidance. 12 research students are successfully awarded M.Phil in Computer Science under his guidance He is also worked on various boides in S.R.T.M.U, Nanded. He is also worked on various bodies is S.R.T.M.U, Nanded. He also published 34 research papers in various international/national journals. He is peer team member of NAAC (National Assessment and Accreditation Council, India). He published a book entitled "DBMS concepts and programming in Foxpro". He also get various educational wards in which "Best Principal"

award from S.R.T.M.U, Nanded in 2009 and "Best Teacher" award from Govt. of Maharashtra, India in 2010. He is life member of Indian "Fellowship of Linnaean Society of London (F.L.S.)" on 11 National Congress, Kolkata (India). He is also honored with November 2009.